\documentclass[12pt]{iopart}

\usepackage{orcidlink}
\usepackage{cite}

\expandafter\let\csname equation*\endcsname\relax
\expandafter\let\csname endequation*\endcsname\relax
\usepackage{amsmath,amssymb,amsfonts}
\usepackage{graphicx}
\usepackage{textcomp}
\usepackage{xcolor}
\usepackage{pdflscape,booktabs,adjustbox}  
\usepackage{makecell} 

\usepackage[T1]{fontenc}
\usepackage{booktabs}
\usepackage{multirow}
\usepackage{amssymb}

\usepackage{algorithm}
\usepackage{algpseudocode}

\usepackage{pifont}

\newcommand{\cmark}{{\color{green!60!black}\ding{51}}} 
\newcommand{\xmark}{{\color{red}\ding{55}}}            
\usepackage{xcolor}

\begin{document}

\title[]{Traces Propagation: Memory-Efficient and Scalable Forward-Only Learning in Spiking Neural Networks}

\author{Lorenzo Pes$^{1,*}$\,\orcidlink{0000-0002-8151-9327}, Bojian Yin\orcidlink{0000-0002-5074-4337}, Sander Stuijk$^1$\,\orcidlink{0000-0002-2518-6847}, Federico Corradi$^1$\,\orcidlink{0000-0002-5868-8077}}

\address{$^1$ Electronic Systems, Eindhoven University of Technology, The Netherlands \\ 
$^*$ Author to whom any correspondence should be addressed.}
\ead{l.pes@tue.nl}
\vspace{10pt}
\textit{Keywords:} spiking neural networks, forward-only, on-device learning, local learning, fine-tuning.

\begin{abstract}
Spiking Neural Networks (SNNs) provide an efficient framework for processing dynamic spatio-temporal signals and for investigating the learning principles underlying biological neural systems. A key challenge in training SNNs is to solve both spatial and temporal credit assignment. The dominant approach for training SNNs is Backpropagation Through Time (BPTT) with surrogate gradients. However, BPTT is in stark contrast with the spatial and temporal locality observed in biological neural systems and leads to high computational and memory demands, limiting efficient training strategies and on-device learning. Although existing local learning rules achieve local temporal credit assignment by leveraging eligibility traces, they fail to address the spatial credit assignment without resorting to auxiliary layer-wise matrices, which increase memory overhead and hinder scalability, especially on embedded devices. In this work, we propose Traces Propagation (TP), a forward-only, memory-efficient, scalable, and fully local learning rule that combines eligibility traces with a layer-wise contrastive loss without requiring auxiliary layer-wise matrices. TP outperforms other fully local learning rules on NMNIST and SHD datasets. On more complex datasets such as DVS-GESTURE and DVS-CIFAR10, TP showcases competitive performance and scales effectively to deeper SNN architectures such as VGG-9, while providing favorable memory scaling compared to prior fully local scalable rules, for datasets with a significant number of classes. Finally, we show that TP is well suited for practical fine-tuning tasks, such as keyword spotting on the Google Speech Commands dataset, thus paving the way for efficient learning at the edge. 


\end{abstract}

\section{Introduction}
One of the fundamental quests of bio-inspired machine intelligence is to formalize learning rules that match the efficiency observed in the biological brain. Recently, spiking neural networks (SNNs) have emerged as a computational model that mimics the efficient behavior of biological neurons, offering a promising substrate for exploring neural system's learning theories. Fundamentally, learning in SNNs entails solving both temporal and spatial credit assignment problems~\cite{2020_lillicrap,Bellec_2020}. The former refers to attributing credit to past synaptic activity based on the effect on future output, while the latter refers to determining how local activity affects higher levels of the network hierarchy. These problems have been successfully addressed by Back-Propagation-Through-Time (BPTT), leading to state-of-the-art performance in SNNs~\cite{Eshraghian_2023}. 

Despite its success, BPTT is widely regarded as biologically implausible because it contradicts several fundamental principles of learning observed in biological neural systems. From a spatial perspective, it suffers from \textit{weight-transport}~\cite{2017_czarnecki} and \textit{update-locking}~\cite{1987_grossberg}, due to its reliance on the chain rule to distribute a global error across the hierarchy. These problems entail a global synchronization mechanism, require signals to traverse synapses in reverse order, and enforce non-local, sequential update, contradicting the highly parallel and local principles of neural systems. From a temporal perspective, BPTT requires knowledge of all future states of the network, violating the time locality observed in biological systems and incurring a memory cost that scales with the sequence length~\cite{2002_werbos}. 

To address these limitations, several learning algorithms have been proposed with the aim of improving biological plausibility and computational efficiency. To mitigate the limitation of non-local temporal credit assignment, three-factor learning rules such as E-prop~\cite{Bellec_2020}, OTTT~\cite{xiao_2022}, OSTL~\cite{ortner_2022}, and S-TLLR~\cite{apolinario_2023} employ eligibility traces, a short-term memory of neural activity, that enables time-local updates. However, they only partially address the limitations of spatial credit assignment. To address both challenges, learning rules such as ETLP~\cite{quintana_2024}, OSTTP~\cite{ortner_2023}, DECOLLE~\cite{kaiser_2020}, and TESS~\cite{apolinario_2025} combine eligibility traces with DRTP~\cite{frenkel_2021} or layer-wise local error losses. Among these, only TESS has been demonstrated in deeper architectures, such as VGG-9, thus paving the way for fully local and scalable learning in SNNs. 

Despite its novelty and competitive performance, TESS requires the storage and transposition of auxiliary matrices for each layer to compute the local loss that addresses the spatial credit assignment problem. These matrices scale with the number of neurons in each layer and with the number of output neurons, leading to significant storage requirements, in particular in convolutional networks, where the number of neurons per layer dominates the memory footprint. Recently, a novel class of learning algorithms, known as Forward-Only \cite{2022_hinton,2023_kohan,2022_dellaferrera}, has emerged as a potential solution to the spatial credit assignment problem, without incurring the cost of layer-wise auxiliary matrices. Among these, Signal Propagation (SP)~\cite{2023_kohan} stands out as the only solution with proven scalability to more complex architectures such as VGG-9. However, its applicability in SNNs remains an open question. To address this, we propose a novel temporally and spatially local learning rule that couples eligibility traces with the SP algorithm, and demonstrate its effectiveness in deep spiking architectures. The main contributions of this work are summarized as follows:

\begin{itemize}
    \item We propose Traces Propagation (TP), a novel and scalable fully-local learning rule that couples the biologically plausible mechanism of eligibility traces for temporal credit assignment with a contrastive loss inspired by Signal Propagation for spatial credit assignment.  

    \item Unlike prior local learning rules, TP does not rely on auxiliary layer-wise projection matrices to perform spatial credit assignment, offering a favorable memory cost for deep architectures with a large number of output classes compared to other state-of-the-art rules.

    \item TP outperforms all other fully local learning rules on the N-MNIST and SHD datasets, and shows competitive performance on DVS-GESTURE and DVS-CIFAR10 using deep architectures such as VGG-9. Furthermore, we demonstrate its suitability for real-world scenarios through fine-tuning on the Google Speech Commands dataset for keyword spotting, highlighting its applicability to resource-constrained edge devices.

\end{itemize}


\subsection{Notation}

Throughout this work, we use the Einstein summation notation with indices enclosed in square brackets (e.g., $[i]$, $[j]$), where repeated indices imply summation over their corresponding dimensions. This convention applies to tensors, and when no index is shown, a scalar quantity is assumed.


\subsection{Background \& state-of-the-art}
\label{sec:bg_sota}

\subsubsection{Neuron Model}
\label{sec:bg_sota:neuron_model}

Throughout this work, we employ the Leaky-Integrate-and-Fire (LIF)~\cite{burkitt_2006}, the most common neuron model used to implement SNNs. In discrete time, the neuronal dynamic of the LIF model is governed by the following equations:

\begin{equation}
    v^t_l[j] = \alpha^l v^{t-1}_l[j] + s^t_{l-1}[i] W_l[i,j] + s^{t-1}_{l}[j] R_l[j,j]- s_{l}^{t-1}[j] v_{th},
\label{eq:lif_mem}
\end{equation}

\begin{equation}
        s^t_l[j] = \Theta (v^t_l[j] - v_{th}).
\label{eq:lif_spike}
\end{equation}

\noindent
In Eq.\ref{eq:lif_mem}, $v^t_l[j]$, represents the membrane potential of neurons in layer $l$ at time step $t$, $W_l[i,j] \in \mathbb{R}^{H_{l-1} \times H_l} $ the synaptic connections between neurons in layer $l$ and $l-1$, $R_l[j,j] \in \mathbb{R}^{H_l \times H_l}$ the recurrent matrix, $\alpha_l = e^{-\Delta t / \tau_{m}}$ the membrane decay with time constant $\tau_{m}$ and $s^t_l$ the spike event. In Eq.\ref{eq:lif_spike}, $v_{th}$ represents the membrane threshold and $\Theta$ the heavy-side step activation function. Due to the discontinuity of the heavy-side step function, a surrogate gradient $\partial  s_l^t[j]/ \partial v_l^t[j] =\Theta'$ is used to enable gradient-based learning \cite{Eshraghian_2023}. 

\subsection{Error Back-Propagation-Through-Time \& its limitations}
\label{sec:bg_sota:bptt_and_lim}

From a machine learning perspective, the problem of learning has been addressed using error back-propagation-through-time (BPTT), leading to state-of-the-art (SOTA) performance in various tasks of different complexity. 
The main objective of BPTT is to provide a method to change the learnable parameters $\theta_l = \{ W_l[i,j], R_l[j,j] \}$, such that the error at the output layer $E_L$ is minimized. This error is calculated by comparing the activity of neurons in the last layer with an expected activity $c^t$. To minimize this loss, the learnable parameters are generally updated following the gradient descent as 

\begin{equation}
    \Delta \theta_l[i,j] = - \eta \cdot \frac{\partial E_L}{\partial \theta_l[i,j]}
\end{equation}

For a network of LIF neurons without recurrent synaptic connections, i.e. $\theta_l = \{ W_l[i,j] \}$, there is still a temporal dependency across time steps due to the leaky integration mechanism of the membrane potential. Specifically, the membrane potential $v_l^t[j]$ at time $t$ depends on its value in the previous time step $v_l^{t-1}[j]$, through the decaying factor $\alpha^l$. Thus, the gradient $\frac{\partial E_L}{\partial \theta_l}$ must account for the influence of each parameter on the entire temporal sequence. To find this gradient, the network is unfolded across the full sequence and BPTT is applied as 
\begin{align}
    \left(\frac{\partial E_L}{\partial 
    \theta_l[i,j]} \right)_{\text{BPTT}} &= \sum^T_{t=1} \frac{\partial E_L}{\partial v_l^t[j]} \frac{\partial v_l^t[j]}{\partial W_l[i,j]} \\
    &= \sum^T_{t=1} 
        \left( 
        \underbrace{
        \frac{\partial E_L}{\partial s_l^t[j]} \frac{\partial s_l^t[j]}{\partial v_l^t[j]}
        }_{\text{spatial}} 
        + 
        \underbrace{
        \frac{\partial E_L}{\partial v_l^{t+1}[j]} \frac{\partial v_l^{t+1}[j]}{\partial v_l^{t}[j]}
        }_{\text{temporal}} 
        \right)
        \underbrace{\frac{\partial v_l^t[j]}{\partial W_l[i,j]}}_{\text{spatial}}.
\label{eq:bptt_grad}
\end{align}

\begin{figure}[t!]
    \includegraphics[width=\textwidth]{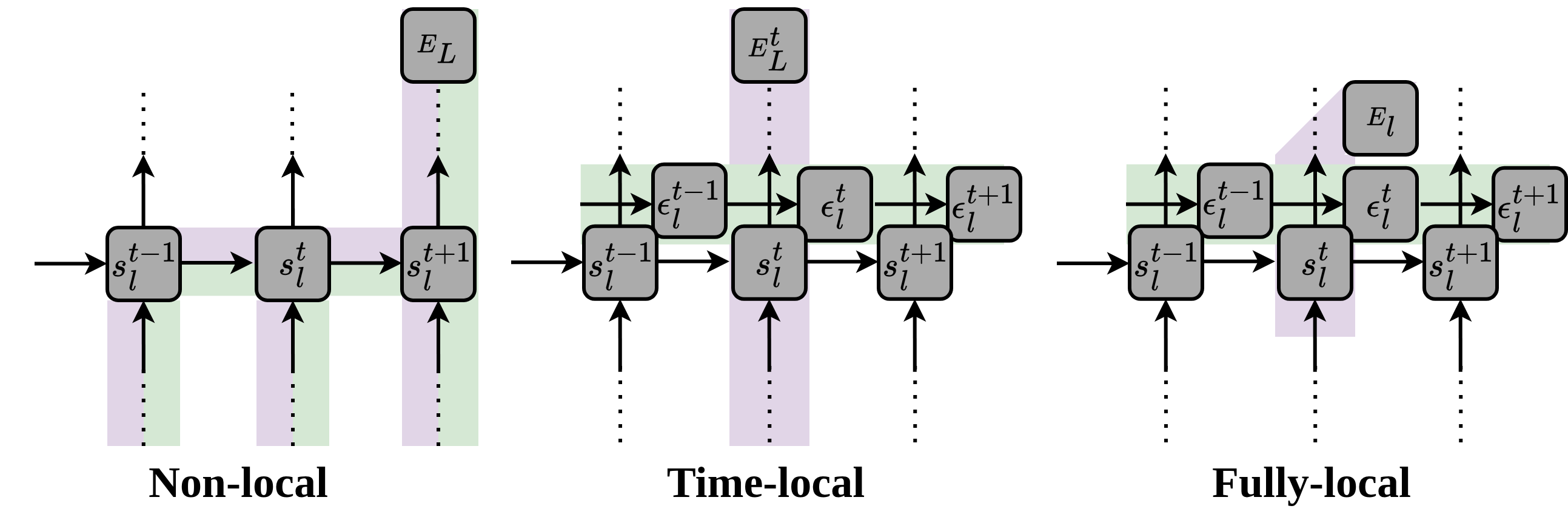}    
       \caption{Locality comparison in learning algorithms. Violet represent the spatial credit assignment, while green indicate the temporal credit assignment. \textit{Left:} Non-local learning (e.g., BPTT) relies on a global error signal $E_L$ that requires both spatial and temporal components for the whole sequence. \textit{Center:} Time-local learning (e.g, E-prop, OSTL, OTTP and S-TLLR) use eligibility traces $\epsilon_l^t$ to remove temporal dependencies, but still requires a global spatial credit assignment based on $E_l^t$. \textit{Right:} Fully-local learning (e.g., ETLP, OSTTP,TESS and TP), still uses eligibility traces for temporal credit assignment but uses only local information for the spatial assignment, thus satisfying both temporal and spatial locality.}
    \label{fig:learning_rules_locality}
\end{figure}

From this equation, it is evident that BPTT consists of both a temporal and a spatial component, which are entangled together, as shown in Fig.~\ref{fig:learning_rules_locality}a. This entanglement exists because the gradient at each time step recursively depends on the future gradient, making the evaluation of the spatial component dependent on the full temporal history. This non-locality of updates leads to a set of problems, which call into question the bio-plausibility of the algorithm and hinder efficient hardware implementations. From a \textbf{spatial} perspective, the main limitations are:
\begin{itemize}
    \item \textbf{Update-locking \cite{2017_czarnecki}}: This issue stems from the fact that the gradient $\partial E_L /\ \partial v_l^t[j]$ in Eq.\ref{eq:bptt_grad} is not available until the full forward pass has been completed and the error of the output layer has been propagated backward through the network. This sequential dependency not only leads to idle resources but also presupposes a global synchronization mechanism that collides with the massively parallel and local processing observed in the biological brain. Furthermore, update-locking entails storing each layer's activation for a given time step until the corresponding gradient can be computed, resulting in significant memory overhead, particularly in deep convolutional networks, where activations dominates the memory footprint.

    \item \textbf{Weight-transport \cite{1987_grossberg}:} In Eq.\ref{eq:bptt_grad}, computing $\partial E_L  /\  \partial v_l^t[j]$ requires multiplying the backpropagated error ($\partial E_L /\ \partial v_{l+1}^t)$ by the transpose of the forward weight matrix:

    \begin{align}
    \frac{\partial E_L}{\partial v_l^t[j]} &=  \frac{\partial E_L}{\partial v_{l+1}^t[k]} \cdot \frac{\partial v_{l+1}^t[k]}{\partial s_l^t[j]} \cdot \frac{\partial s_l^t[j]}{\partial v_l^t[j]}
    \label{eq:wt_full}
    \\
    &=  \frac{\partial E_L}{\partial v_{l+1}^t[k]} \cdot W_{l+1}[j,k] \cdot \frac{\partial s_l^t[j]}{\partial v_l^t[j]}
    \label{eq:wt_explicit}
    \end{align}

    This implies that the output error must traverse the same network in the reverse direction to perform the synaptic update, a principle that is hard to justify from a biological perspective~\cite{lillicrap_2016}. Furthermore, transposing matrices in digital systems leads to non-contiguous memory accesses, which increases latency and energy consumption \cite{2004_tsifakis,2019_lopes}.
    
\end{itemize}

\noindent
From a \textbf{temporal} perspective, the primary limitation arises from the inherent temporal recurrence of the membrane potential, which results in:

\begin{itemize}
    \item \textbf{Activations history}: As expressed in Eq.~\ref{eq:bptt_grad}, the gradient $\partial E_L /\ \partial v_l^t[j]$ recursively depends on future terms via $\partial v_l^{t+1}[k] /\ \partial v_l^t[j]$. This dependency requires access to the entire sequence of membrane potential states, which must be stored during the forward pass to compute gradients during backpropagation. This effect imposes a memory footprint that scales linearly with the length of the temporal sequence ($T$), the number of neurons in each layer ($H$) and the number of layers ($L$), resulting in a spatial complexity of $\mathcal{O}(THL)$. Furthermore, this non-local dependency collides with the local update mechanism observed on the brain, where each synaptic update depends exclusively on the current or recent states of neurons. 
\end{itemize}

These challenges not only highlight the biological implausibility of BPTT but also limit its implementation in resource-constrained hardware accelerators. In the following sections, we will introduce how the limitations of spatial and temporal credit assignment of BPTT have been addressed in the literature. 



\subsubsection{Addressing Limitations in Temporal Credit Assignment}
\label{sec:bg_sota:address_time}

To solve the non-locality of the temporal credit assignment in BPTT, \textit{three-factor} learning rules have been proposed \cite{zenke_2018, ortner_2022, xiao_2022, apolinario_2023, Bellec_2020}. This family of learning algorithms approximate the BPTT gradients as

\begin{equation}
          \left(\frac{\partial E_L}{\partial 
\theta_l [i,j]} \right)_{\text{BPTT}} \approx  \left(\frac{\partial E_L}{\partial \theta_l [i,j]} \right)_{\text{three-factors}} = \sum_t S_l^t[j] \epsilon_l^t[i,j]
\end{equation}

\noindent
where $S_l^t[j]$ is a top-down learning signal that acts as a modulatory factor, conveying spatial credit assignment information. In contrast, $\epsilon_l^t[i,j]$ is a transient synaptic variable known as the \textit{eligibility trace}, supported by experimental findings in biological neurons \cite{gerstner_2018}. It captures the history of synaptic activity through an exponential low-pass filter version of both pre- and post-synaptic activities, and locally encodes the \textit{temporal credit assignment}. This separation allows learning to occur in a time-local and biologically plausible manner, as illustrated in Fig.~\ref{fig:learning_rules_locality}. In generic form, eligibility traces are defined as 

\begin{equation}
\epsilon_l^t[i,j] = \beta \epsilon_l^{t-1}[i,j] + g(s_{l-1}^t[i])f(s_{l}^t[j])
\label{eq:trace_generic},
\end{equation}

\noindent
 where $\beta$ is a decay factor. The precise formulation of $g(\cdot)$, $f(\cdot)$ and the learning signal $S_l^t[j]$ vary between different rules depending on how the spatio-temporal gradients are derived and separated. For instance, the eligibility trace of E-prop \cite{Bellec_2020} defines the presynaptic factor as a low-pass filtered version of the spiking activity, $g(s_{l-1}^t[i]) = \alpha \, g(s_{l-1}^{t-1}[i]) + s_{l-1}^t[i]$, 
and the postsynaptic factor as the surrogate derivative of the spike function, 
$f(s_l^t[j]) = \Theta' (v^{t+1}_l[j])$. The learning signal is defined as an output-based modulatory signal as $(S_l^t[j])_{\text{e-prop}} = \sum_{k=1}^C(c^t_L[k]-c[k])B_l[k,j]$, with C the number of output neurons and $B_l$ providing a fixed mapping that implements a structural linear dimensionality reduction from layer $l$ to the output layer. 

On the other hand, OSTL \cite{ortner_2022}, provides a more accurate approximation of BPTT where eligibility traces are derived to be equivalent to BPTT for single-layer recurrent networks, while maintaining temporal locality. Moreover, differently from E-prop \cite{Bellec_2020}, OSTL explicitly separates the spatial and temporal components in a layer-wise manner, enabling the learning signal to be the exact spatial gradient of error backpropagation.

While these solutions offer a biologically plausible solution to the non-locality of temporal credit assignment of BPTT, enabling time-local updates without storing activations at each time step, they still suffer from a significant memory overhead. 

Specifically, eligibility traces are stored per synapses, leading to a space complexity that scales quadratically with the number of neurons, i.e., $\mathcal{O}(H^2L)$, making it prohibitive for deep architectures. To reduce the quadratic complexity of standard eligibility formulations, recent learning rules such as S-TLLR \cite{apolinario_2023} and OTTP~\cite{xiao_2022} propose to set $\beta = 0$ in Eq.\ref{eq:trace_generic}, and keep only filtered versions of the pre- and post-synaptic spiking activities. These formulations enable an update rule that depends solely on local activity trading off performance for a space complexity of $\mathcal{O}(HL)$. 

\subsubsection{Addressing Limitations in Spatial Credit Assignment}
\label{sec:bg_sota:adress_space}

To address the limitations of spatial credit assignment, different alternatives have been proposed in the literature, including Feedback-Alignment (FA) \cite{lillicrap_2016}, Direct-Feedback-Alignment (DFA) \cite{nokland_2016}, and Direct-Random-Target-Projection (DRTP) \cite{frenkel_2021}. FA addresses the weight-transport problem by replacing $W^\top$ in Eq.\ref{eq:wt_explicit} with a fixed random matrix $B_l$ of the same dimensions. DFA builds on this idea but proposes to bypass the backward pass by directly projecting the output error gradient to each hidden layer using a fixed random matrix, resulting in the following update rule:

\begin{equation}
    \left( \frac{\partial E_L}{\partial v_l^t[j]} \right)_{\text{DFA}} = \frac{\partial E_L}{\partial v_L^t[k]} \cdot B_l[k,j] \cdot \frac{\partial s_l^t[j]}{\partial v_l^t[j]}
    \label{eq:dfa}
\end{equation}

This strategy is used in OSTL to derive a more biologically plausible learning rule, OSTL(rnd), which maintains temporal locality while eliminating the weight-transport problem. Nevertheless, these solutions still rely on top-down error propagation from the output layer, thereby leaving unaddressed the update-locking problem. 

To address both the update-locking and weight-transport problems, DRTP was proposed in~\cite{frenkel_2021} as a purely feedforward alternative, where the target output ($c^*$) replaces the gradient term \( \partial E_L / \partial v_L^t \) in Eq.~\ref{eq:dfa} and is directly projected to the hidden layers. This eliminates the need to wait for a top-down error propagation, as the target vector is available before the forward pass begins. This approach not only enables the pipelining of input signals, but also eliminates the need to store activations at each time step.

Nonetheless, all these approaches require an additional projection matrix for each hidden layer (i.e. $B_l$). While this can be sustainable in terms of storage for multilayer perceptrons, it can lead to significant memory overhead in convolutional networks, where the activations dominate the memory footprint. Thus, with the intention of removing both the backward pass and the storage overhead of auxiliary projection matrices, a novel class of learning algorithms known as \textit{Forward-Only} \cite{2022_dellaferrera, 2022_hinton, 2023_kohan} has recently emerged. 

Forward-Only algorithms fully eliminate the backward pass and rely exclusively on multiple forward passes to drive learning. For example, PEPITA~\cite{2022_dellaferrera} drives its learning based solely on the activity difference between two forward passes:\mbox{ (1)} a standard forward propagation of the input signal and (2) a forward propagation of an error-modulated version of the same input. On the other hand, Forward-Forward \cite{2022_hinton} and Signal Propagation \cite{2023_kohan} are inspired by contrastive learning to drive the parameter updates. For example, Forward-Forward optimizes each layer parameter to distinguish between \textit{positive} (real) and \textit{negative} (fake) inputs by maximizing a local \textit{goodness} metric, generally the squared sum of activities, for positive examples and minimizing it for negative ones. Specifically, positive examples are input data paired with correct labels, while negative examples are corrupted input data paired with incorrect labels. 

In contrast, SP \cite{2023_kohan} minimizes a layer-wise cross-entropy loss computed over pairwise similarities between input activations and target activation within the same batch. These targets are generated by projecting the desired batched output targets $c^*$, through a target propagator that brings the targets to the same dimension as the first hidden layer, allowing direct comparison. Thus, by minimizing the local loss, SP maximizes the similarity between an input example activation and the target of its corresponding class, while minimizing the similarity to targets of other classes within the same batch. Furthermore, during training, the targets are also jointly optimized to move closer to the representation of their respective class and further from those of other classes, thereby generating linearly separable representations.  

Notably, SP and PEPITA require only a single additional projection matrix, used for the target propagator and for the error-based modulation factor, respectively, whereas Forward-Forward require a mechanism to construct positive and negative examples. These approaches significantly reduce the memory footprint compared to previously discussed algorithms, as no additional matrices are needed to scale to deeper architectures. Nonetheless, among Forward-Only algorithms, only Signal Propagation has been demonstrated to scale effectively to deeper networks, making it a suitable candidate to solve the limitations of spatial credit assignment in deep architectures.

\subsubsection{Towards fully-local learning in SNNs}
\label{sec:bg_sota:fully_local}

Research for bio-inspired and hardware-efficient learning rules for training SNNs is increasingly directed toward fully local learning, as depicted in Fig.\ref{fig:learning_rules_locality}c. These kinds of learning rules attempt to approximate gradient-based learning using only information that is locally available in time and space, thus eliminating the need for global signals, weight-transport, and storage of temporal activity. One of the first examples of this kind of learning rules is DECOLLE, proposed in \cite{kaiser_2020}. It employs random readout layers that project the spiking activations $s_l^t$ of each hidden layer onto the dimension of the true target vector $c^*$, through random fixed matrices $B_l \in \mathbb{R}^{H_l \times C}$, to generate local targets $c_l^t$. In each layer a local Mean Squared Error (MSE) is computed between $c_l^t$ and $c^*$, and the gradient of this loss with respect to the learnable parameters provides a post-synaptic learning signal. This signal is then multiplied with a second-order low-pass filtered version of the pre-synaptic activity, forming a local three-factor learning rule that enables fully local, online updates. 

Algorithms such as ETLP \cite{quintana_2024} and OSTTP \cite{ortner_2023} both employ the learning signal proposed in DRTP as a learning signal to modulate eligibility traces. Specifically, ETLP employs the eligibility traces defined in e-prop \cite{Bellec_2020}, while OSTTP uses the trace formulation of OSTL \cite{ortner_2022}. Nevertheless, all the aforementioned solutions have been demonstrated only for small-scale architecture. Among them, DECOLLE stands out as the only method demonstrated on a network with up to three convolutional layers, highlighting its relative scalability in the context of deep SNNs.

Recently, a novel learning algorithm known as TESS \cite{apolinario_2025} has been proposed, being the first to showcase fully local learning in deep structures, scaling up to VGG-9 architectures. It builds on the same eligibility traces of \cite{apolinario_2023}, but introduces a novel local learning mechanism for spatial credit assignment. It works similarly to the local loss proposed by DECOLLE, but instead of computing the gradient of the MSE loss, it projects back a direct comparison between a generated target and the true target through the transposition of matrix B, resulting in a learning signal $S_l^t[j] = B_l^\top \left( f\left( B_l[c,j] \cdot s_l^t[j] \right) - c^*[j] \right)$.

Despite their differences, all the aforementioned algorithms require additional projection matrices (e.g., $B_l$) at each layer to solve the spatial credit assignment problem. Although this is effective and shows scalability, as in the case of TESS, it introduces a significant memory overhead, particularly in convolutional architectures where such matrices must match the number of channels and spatial dimensions. In contrast, leveraging forward-only strategies such as Signal Propagation discussed in Section~\ref{sec:bg_sota:adress_space} offers a promising alternative where only one extra projection matrix is required for the first layer. 

\section{Results}
\label{sec:result}

The application of Signal Propagation \cite{2023_kohan} to SNNs remains unclear due to the inherent temporal dynamic of these architectures. In this work, we hypothesize that the use of eligibility traces is sufficient to learn temporal dynamics using the Signal Propagation algorithm. Inspired by its formalism, our proposed solutions aim at clustering the layer-wise activity traces of the input examples belonging to the same class close to the target traces of the same class examples and apart from target traces of examples belonging to other classes. To derive the update rule of the Traces Propagation algorithm in the following section, we employ the same LIF model as defined in Eq.~\ref{eq:lif_mem} and Eq.~\ref{eq:lif_spike}, and follow the previously introduced notation and assumed dimensionality. 

In this section, we begin by explaining the update rule of Traces Propagation in Section~\ref{sec:result:traces_propagation}. Following this introduction to mathematical theory, we present experimental results on conventional SNN benchmarks in Section~\ref{sec:result:experiemntal_results} and a comparison for hardware implementation in Section~\ref{sec:result:hw_comp}. Ultimately, we provide real-world use cases for fine-tuning in Section~\ref{sec:result:use_cases}.

\subsection{Traces Propagation}
\label{sec:result:traces_propagation}

\begin{figure}[t!]
    \centering
    \includegraphics[width=0.5\textwidth]{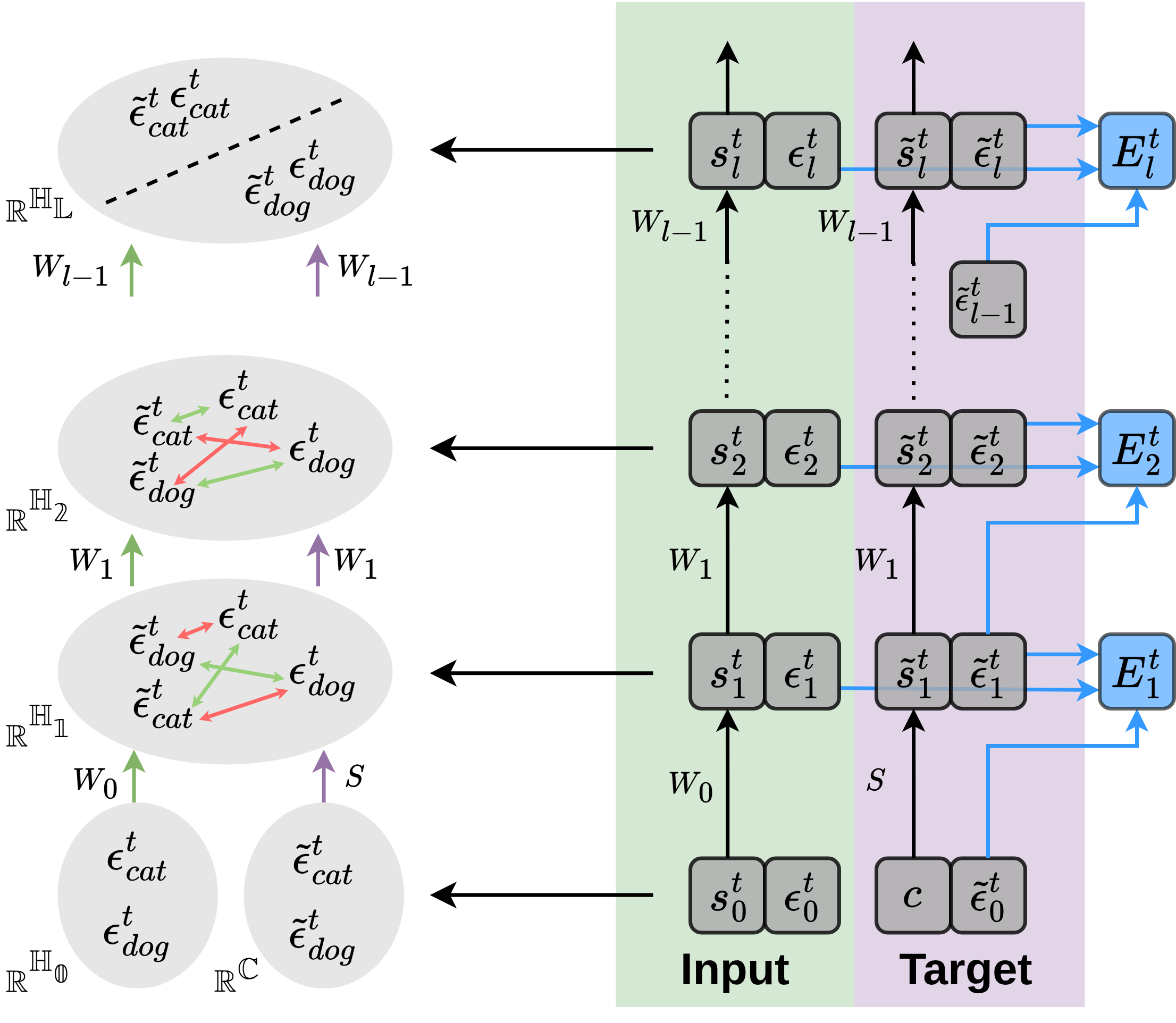}    
       \caption{Overview of Traces Propagation. \textit{Right}: Parallel computational paths. In the green path, the input signal $s_0^t$ is propagated through the network, via conventional matrices $W_l$, generating input traces $\epsilon_l^t$ based on spiking activity $s_l^t$. In the purple path, the one-hot encoded target vector $c \in \mathcal{R^C}$ is projected to the first layer via $S \in \mathcal{R}^{C \times H_{1}}$, to match the dimensionality of the first input trace (i.e. $\epsilon_1^t$), thus enabling the comparison of input and target traces through local loss $E_l^t$. For $l>1$, the same matrices $W_l$ are used to propagate the target signal and generate target traces $\tilde{\epsilon}^t_l$ at each layer. \textit{Left}: Geometric view and trace alignment dynamics. Target traces attracts input traces of the same class (green arrows) and repel those from different classes (red arrows), fostering class separation. At the last layers, the input traces of different classes become linearly separable (dashed line).}
    \label{fig:trace_prop}
\end{figure}




At each layer $l$, we maintain two distinct activity traces, the input and target traces, computed as a leaky integration of the local spiking activity defined as

\begin{align}
  \epsilon_{l}^{t}[b,j]          &= \beta_{l}\,\epsilon_{l}^{t-1}[b,j] + s_{l}^{t}[b,j],          
  \label{eq:trace_in} \\
  \tilde{\epsilon}_{l}^{t}[b,j] &= \beta_{l}\,\tilde{\epsilon}_{l}^{t-1}[b,j] + \tilde{s}_{l}^{t}[b,j], 
  \label{eq:trace_trg}
\end{align}

\noindent
where \eqref{eq:trace_in} and \eqref{eq:trace_trg} represent the trace obtained by propagating through the network the input spikes $s_0^t$ and the one-hot encoded target vector $c$, respectively. In particular, in this context, we make the batch index $b$ explicit, as it will play a fundamental role in the loss definition. For clarity, we will refer to \eqref{eq:trace_in} as the \textit{input trace}, and to \eqref{eq:trace_trg} as the \textit{target trace}. At each layer, the local loss is inspired by \cite{2023_kohan} and defined as 

\begin{equation}
    E^t_l = -y^t_{l}[b,b'] \log\left(\text{Softmax}(z^t_{l}[b,b'])\right)
    \label{eq:tp_loss}
\end{equation}

\noindent
where $z_l^t$ and $y_l^t$ are defined as

\begin{equation}
   z_l^t[b,b'] = \epsilon^t_l[b,j] \tilde{\epsilon}^t_l[j,b'],
   \label{eq:act_comp}
\end{equation}


\begin{equation}
   y_l^t[b,b'] = \text{Softmax}\left( f\left(\tilde{\epsilon}^t_{l-1}[b,j],\tilde{\epsilon}^t_{l-1}[j,b']\right) \right).
   \label{eq:act_comp_prev}
\end{equation}

Equation \eqref{eq:act_comp} are the logits of the CE loss and is a dot-product similarity score, computed over the feature dimension, that measures how well the trace of the input example $b$ aligns with the target trace of the corresponding example $b'$. On the other hand, \eqref{eq:act_comp_prev} are the targets of the CE loss, defined as a row-wise softmax of a generic similarity function $f(\cdot)$ (e.g dot product or negative euclidean distance), computed between the examples of the target traces of the previous layer. In the first layer, the target traces are derived from one-hot encoded class labels, which makes them linearly separable. Thus, for the first layer, \eqref{eq:act_comp_prev} defines a probability distribution that places a high probability mass on examples belonging to the same class. As the target traces are propagated through the network, more abstract features are generated, while preserving their semantic structure, such that examples from the same class remain close in the target space. Consequently, by minimizing the loss \eqref{eq:tp_loss} in each layer with respect to the learnable parameters $\theta_l$ as per Eq.~\eqref{eq:chan_rule_init}, the networks learn to move the traces of each sample close to their target traces while pushing them away from the representations of other samples, resulting in linearly separable representations at the final layer, as shown in Fig.~\ref{fig:trace_prop}.  

\begin{align}
\frac{\partial E^t_l}{\partial \theta_{l}[i,j]} &= \frac{\partial E^t_{l}}{\partial z_l^t}\frac{\partial z_l^t}{\partial \theta_l}\label{eq:chan_rule_init} \\ &= \frac{\partial E^t_{l}}{\partial z_l^t}  \left[
\underbrace{
\frac{\partial z_l^t}{\partial \epsilon_l^t}
\frac{\partial \epsilon_l^t}{\partial s_l^t}
\frac{\partial s_l^t}{\partial v_l^t}
\frac{\partial v_l^t}{\partial \theta_{l}}
}_{\text{input}} 
+ 
\underbrace{
\frac{\partial z_l^t}{\partial \tilde{\epsilon}_l^t} 
\frac{\partial \tilde{\epsilon}_l^t}{\partial \tilde{s}_l^t}
\frac{\partial \tilde{s}_l^t}{\partial \tilde{v}_l^t}
\frac{\partial \tilde{v}_l^t}{\partial \theta_{l}}
}_{\text{target}}
\right]
\label{eq:chain_rule_grad} \\
&= (Softmax(z_l^t) - y_l^t)[b,b'] \Big[ \,
\tilde{\epsilon}_l^t[b',j] \, \Theta'(v_l^t[b,j] - V_{\text{th}}) \, s^t_{l-1}[b,i] \notag \\
&\hspace{2.4cm} 
+ \, \epsilon_l^t[b,j] \, \Theta'(\tilde{v}_l^t[b',j] - V_{\text{th}}) \, \tilde{s}^t_{l-1}[b',i] 
\, \Big]
\label{eq:local_grad}
\end{align}

In \eqref{eq:local_grad} the signal $(Softmax(z_l^t) - y_l^t)[b,b']$ is a modulatory signal that measures how much the input trace of sample $b$ and the target trace of sample $b'$ should be pulled together or pushed apart, thus creating class-based clustering. This push and pull interaction is depicted by the green and red arrows, respectively, on the right side of Fig.\ref{fig:trace_prop}. Notably, each synaptic update, follows a typical three-factor rule, as explained in Section~\ref{sec:bg_sota:address_time}, and requires information that is local in time and space. However, due to its reliance on inter-sample contrast (i.e. $[b,b']$), this update rule cannot be computed in an online, per-sample fashion and requires a batch size of at least 2. We provide the algorithm for the proposed update rule in~\ref{appendix:1}.

\subsection{Experimental Results}
\begin{table}[t!]
    \centering
    \caption{Comparison to SoTA results on N-MNIST and SHD}
    \label{tab:fully_connected}
    \setlength{\tabcolsep}{3pt}
    \resizebox{\columnwidth}{!}{
    \begin{tabular}{c c c c c c c c}
        \toprule
        \textbf{Model} & \makecell{\textbf{Architecture} \\ \textbf{Type}} & \makecell{\textbf{Neuron} \\ \textbf{Type}} & \makecell{\textbf{Number of} \\ \textbf{Neurons}} & \makecell{\textbf{Local} \\ \textbf{Learning}} & \makecell{\textbf{Time} \\ \textbf{Steps}} & \makecell{\textbf{Number of} \\ \textbf{Epochs}} & \makecell{\textbf{Test} \\ \textbf{Accuracy}} \\
        \midrule
        \multicolumn{8}{c}{\textbf{N-MNIST}} \\
        \midrule
        BPTT (ours)            & Feed-Forward & LIF  & 200 & \xmark            & 10  & 100 & 98.45 $\pm$ 0.04 \\
        eProp~\cite{Bellec_2020}              & Feed-Forward & LIF  & 200 & Partial (time)    & 10  & 100 & 97.90$^1$ \\
        ETLP~\cite{quintana_2024}               & Feed-Forward & LIF  & 200 & \cmark            & 10  & 100 & 94.30 \\
        DECOLLE~\cite{kaiser_2020}            & Feed-Forward & LIF  & 200 & \cmark            & 10  & 100 & 96.27$^1$ \\
        \textbf{TP (ours)}     & \textbf{Feed-Forward} & \textbf{LIF}  & \textbf{200} & \textbf{\cmark} & \textbf{10} & \textbf{100} & \textbf{97.33} $\pm$ \textbf{0.06} \\
        \midrule
        \multicolumn{8}{c}{\textbf{SHD}} \\
        \midrule
        BPTT (ours)            & Feed-Forward & LIF  & 450 & \xmark            & 100 & 100 & 75.85 $\pm$ 0.48 \\
        eProp~\cite{Bellec_2020}              & Feed-Forward & LIF  & 450 & Partial (time)    & 100 & 100 & 63.04$^1$ \\
        ETLP~\cite{quintana_2024}               & Feed-Forward & ALIF & 450 & \cmark            & 100 & 100 & 59.19 \\
        DECOLLE~\cite{kaiser_2020}            & Feed-Forward & LIF  & 450 & \cmark            & 100 & 100 & 58.55$^1$ \\
        \textbf{TP (ours)}     & \textbf{Feed-Forward} & \textbf{LIF}  & \textbf{400} & \textbf{\cmark} & \textbf{100} & \textbf{100} & \textbf{67.06} $\pm$ \textbf{0.96} \\
        \midrule
        BPTT (ours)            & Recurrent    & LIF  & 450 & \xmark            & 100 & 100 & 83.23 $\pm$ 1.00 \\
        S-TLLR~\cite{apolinario_2023}                 & Recurrent    & LIF  & 450 & Partial (time)    & 100 & 100 & 78.24 $\pm$ 1.84 \\
        eProp~\cite{Bellec_2020}              & Recurrent    & LIF  & 450 & Partial (time)    & 100 & 100 & 80.79$^1$ \\
        OSTTP~\cite{ortner_2023}                  & Recurrent    & SNU  & 450 & \cmark            & 100 & 100 & 77.33 $\pm$ 0.80 \\
        ETLP~\cite{quintana_2024}               & Recurrent    & ALIF & 450 & \cmark            & 100 & 100 & 74.59$^1$ \\
        DECOLLE~\cite{kaiser_2020}            & Recurrent    & LIF  & 450 & \cmark            & 100 & 100 & 62.01$^1$ \\
        \textbf{TP (ours)}     & \textbf{Recurrent} & \textbf{LIF}  & \textbf{450} & \textbf{\cmark} & \textbf{100} & \textbf{100} & \textbf{81.80} $\pm$ \textbf{0.51} \\
        \bottomrule
    \end{tabular}
    }
    \vspace{2mm}
    \raggedright
    \scalebox{0.8}{%
      \parbox{\linewidth}{%
        $^1$ Results from \cite{quintana_2024}.
      }%
    }
\end{table}

\begin{table}[t!]
    \centering
    \caption{Comparison to SoTA results on IBM-GESTURE and DVS-CIFAR10}
    \label{tab:convolutional}
    \setlength{\tabcolsep}{3pt}
    \resizebox{\columnwidth}{!}{
    \begin{tabular}{c c c c c c c}
        \toprule
        \makecell{\textbf{Model}} & \makecell{\textbf{Architecture} \\ \textbf{Type}} & \makecell{\textbf{Neuron}\\\textbf{Type}} & \makecell{\textbf{Local} \\ \textbf{Learning}} & \makecell{\textbf{Time}\\\textbf{Steps}} & \makecell{\textbf{Number of}\\\textbf{Epochs}} & \makecell{\textbf{Test} \\ \textbf{Accuracy}} \\
        \midrule
        \multicolumn{7}{c}{\textbf{DVS-GESTURE}} \\
        \midrule
        BPTT (ours)            & VGG-9             & LIF           & \xmark & 20   & 200 & 98.56 $\pm$ 0.15  \\     
        OTTT~\cite{xiao_2022}                   & VGG-9             & LIF           & Partial (time) & 20   & 200 & 96.88             \\
        S-TLLR~\cite{apolinario_2023}                 & VGG-9             & LIF           & Partial (time) & 20   & 200 & 98.48 $\pm$ 0.37  \\
        DECOLLE~\cite{kaiser_2020}                & 3-CONV            & LIF           & \cmark & 1800 & 200 & 95.54 $\pm$ 0.16  \\
        \textbf{TESS~\cite{apolinario_2025}}      & \textbf{VGG-9}    & \textbf{LIF}  & \textbf{\cmark} & \textbf{20}   & \textbf{200} & \textbf{98.56} $\pm$ \textbf{0.31} \\
        TP(\textbf{ours})      & VGG-9             & LIF           & \cmark & 20   & 200 & 98.18 $\pm$ 0.15  \\
        \midrule
        \multicolumn{7}{c}{\textbf{DVS-CIFAR10}} \\
        \midrule
        BPTT~\cite{apolinario_2025}           & VGG-9             & LIF           & \xmark & 10   & 200 & 76.40 $\pm$ 0.66  \\
        OTTT~\cite{xiao_2022}                   & VGG-9             & LIF           & Partial (time) & 10   & 200 & 76.27 $\pm$ 0.05  \\
        S-TLLR~\cite{apolinario_2023}                 & VGG-9             & LIF           & Partial (time) & 10   & 200 & 75.14 $\pm$ 1.37  \\
        \textbf{TESS~\cite{apolinario_2025}}      & \textbf{VGG-9}    & \textbf{LIF}  & \textbf{\cmark} & \textbf{10}   & \textbf{200} & \textbf{75.00} $\pm$ \textbf{0.65} \\
        TP(\textbf{ours})      & VGG-9             & LIF           & \cmark & 10   & 200 & 71.39 $\pm$ 0.19  \\
        \bottomrule
    \end{tabular}
    }
\end{table}

\label{sec:result:experiemntal_results}
Following the mathematical formulation, we evaluate the effectiveness of the proposed learning rule on standard datasets for SNNs such as N-MNIST~\cite{orchard_2015}, SHD~\cite{cramer_2020}, DVS-GESTURE~\cite{2017_amir}, and DVS-CIFAR10~\cite{2017_li}. The details of the pre-processing and the hyperparameter tuning strategy are discussed in detail in Section~\ref{sec:material_and_method:setup_and_datasets}. Specifically, we test TP in both fully connected and convolutional architectures: N-MNIST and SHD are evaluated with fully connected architectures, using recurrent connections for the case of SHD, while DVS-GESTURE and DVS-CIFAR10 are tested with VGG-9 convolutional neural networks, to demonstrate its scaling capabilities.

The results for fully connected architectures are reported in Table~\ref{tab:fully_connected}, while those for convolutional architectures are reported in Table~\ref{tab:convolutional}. Remarkably, TP maintains the highest performance across all fully local algorithms for both N-MNIST and SHD. Compared to BPTT, TP loses only 1.12 percentage points (pp) on N-MNIST with feed-forward architectures and 1.43 pp on SHD with recurrent connections. On more complex datasets like DVS-GESTURE and DVS-CIFAR10, TP maintains competitive performance even in deep architectures like VGG-9, but fails to achieve the state-of-the-art performance achieved by TESS, the only other fully local and scalable algorithm. Specifically, TP loses only 0.33 pp compared to TESS and BPTT on the DVS-GESTURE, while on the DVS-CIFAR10 dataset, TP loses 4.78 pp from BPTT 3.38 pp from TESS. We attribute this larger discrepancy to the contrastive nature of TP, which can be hindered by the high inter-class similarity of the CIFAR10 datasets~\cite{2016_murthy}.

\subsection{Memory and Computation Complexity}
\label{sec:result:hw_comp}
\begin{table}[t!]
    \centering
    \caption{Comparison of learning algorithms. The table highlights the key issues that limit efficient hardware implementation such as update-locking, weight-transport, time locality, and space locality. Furthermore, we include the time and space complexity of each algorithm, together with extra storage required for auxiliary matrices (e.g., $B_l$). The variables are as follows: $L$ is the total number of layers, $H$ is the number of neurons, $T$ is the sequence length, and $O$ is the number of output classes.}
    \label{tab:algo_hw_full}
    \setlength{\tabcolsep}{4pt}
    \resizebox{\columnwidth}{!}{
    \begin{tabular}{c c c  c c  c c  c}
        \toprule
        \textbf{Model} & \makecell{\textbf{Update} \\ \textbf{Locking}} & \makecell{\textbf{Weight} \\ \textbf{Transport}} & \makecell{\textbf{Time} \\ \textbf{Local}} & \makecell{\textbf{Space} \\ \textbf{Local}} & \makecell{\textbf{Space} \\ \textbf{Complexity}} & \makecell{\textbf{Time} \\ \textbf{Complexity}} & \makecell{\textbf{Auxiliary} \\ \textbf{Matrices}} \\
        \midrule
        BPTT                                        & \xmark & \xmark & \xmark & \xmark & $TLH$   & $TLH^2$   & $-$    \\
        \midrule
        E-prop \cite{Bellec_2020}                   & \xmark & \xmark & \cmark & \xmark & $LH^2$  & $LH^2$    & $-$    \\
        E-prop(\textit{rnd}) \cite{Bellec_2020}     & \xmark & \cmark & \cmark & \xmark & $LH^2$  & $LOH$     & $LOH$  \\
        OSTL \cite{ortner_2022}                     & \xmark & \xmark & \cmark & \xmark & $LH^2$  & $LH^2$    & $-$    \\
        OSTL(\textit{rnd}) \cite{ortner_2022}       & \xmark & \cmark & \cmark & \xmark & $LH^2$  & $LOH$     & $LOH$  \\
        S-TLLR \cite{apolinario_2023}               & \xmark & \xmark & \cmark & \xmark & $LH$    & $LH^2$    & $-$    \\
        \midrule
        DECOLLE \cite{kaiser_2020}                  & \cmark & \cmark & \cmark & \cmark & $LH$    & $LOH$     & $LOH$  \\
        ETLP \cite{quintana_2024}                   & \cmark & \cmark & \cmark & \cmark & $LH$  & $LOH$     & $LOH$  \\
        OSTTP \cite{ortner_2023}                    & \cmark & \cmark & \cmark & \cmark & $LH^2$  & $LOH$     & $LOH$  \\
        TESS \cite{apolinario_2023}                 & \cmark & \cmark & \cmark & \cmark & $LH$    & $LOH$     & $LOH$  \\
        \textbf{TP (ours)}                          & \cmark & \cmark & \cmark & \cmark & $LH$    & $LH$      & $OH$   \\
        \bottomrule
    \end{tabular}
    }
\end{table}

In this section, we characterize Traces Propagation in terms of its memory and computational complexity, and compare it to BPTT and other local learning rules. The results of this comparison are summarized in Table~\ref{tab:algo_hw_full}. In this analysis, we assume an equal number of neurons $H$ in all layers $L$. Like in \cite{apolinario_2025}, for estimating the time complexity, we exclude the computation time for the traces and the outer product with the presynaptic term, which would otherwise dominate the expression with a $\mathcal{O}(H^2)$ term and make all algorithms appear similarly expensive. Thus, the time complexity is expressed considering only the modulatory signal. In the case of TP, the modulatory signal is $(y_l^t - z_l^t)[b,b']$. The time to compute this modulatory signal is $\mathcal{O}(B^2LH)$, as it is obtained by computing pairwise similarities between all inputs and target traces at each layer for each example in the batch. Compared to other fully-local algorithms such as DECOLLE~\cite{kaiser_2020}, ETLP~\cite{quintana_2024}, OSTTP~\cite{ortner_2023} and TESS\cite{apolinario_2025}, which have a time complexity of $\mathcal{O}(BLOH)$, due to the output error or target projections to generate the modulation signal, TP trades the dependency on the number of output neurons for a factor of $B^2$.

The space complexity at each time step of all time-local learning rules is dominated by the traces storage. For instance, OSTTP~\cite{ortner_2023}, which is based on OSTL~\cite{ortner_2022} stores traces in a per-synaptic fashion, leading to a space complexity of $\mathcal{O}(LH^2)$. On the other hand, algorithms such as ETLP\cite{quintana_2024}, S-TLLR~\cite{apolinario_2023}, DECOLLE~\cite{kaiser_2020} and TESS~\cite{apolinario_2025}, store traces per neuron, leading to a reduced space complexity of $\mathcal{O}(LH)$. Similarly, as defined in Equations \ref{eq:trace_in} and \ref{eq:trace_trg}, TP requires two traces that are stored per neuron, one for the input signal and another for the target signal, leading to a total memory cost of $2BLH$ and a complexity of $\mathcal{O}(LH)$. However, while TP has a comparable space complexity to other fully-local algorithms, as mentioned in Section~\ref{sec:bg_sota:fully_local}, these algorithms require storing auxiliary matrices $B_l$ to perform the spatial credit assignment, resulting in an additional memory cost of $LOH$. On the other hand, TP requires only a single auxiliary matrix for the target propagator, reducing the memory cost to $OH$. A detailed explanation of the computation and memory requirements for TP is provided in Section~\ref{sec:material_and_method:mac_and_mem}. 


\subsection{Use cases}
In this section, we further evaluate TP in a real-world use case. Specifically, we demonstrate its ability to perform efficient on-device fine-tuning for use-specific adaptation in speech recognition.
\label{sec:result:use_cases}

\subsubsection{Fine-Tuning on Google Speech Commands}
\label{sec:result:use_cases:gsc_fine_tune}

\begin{figure}[t!]
    \includegraphics[width=\textwidth]{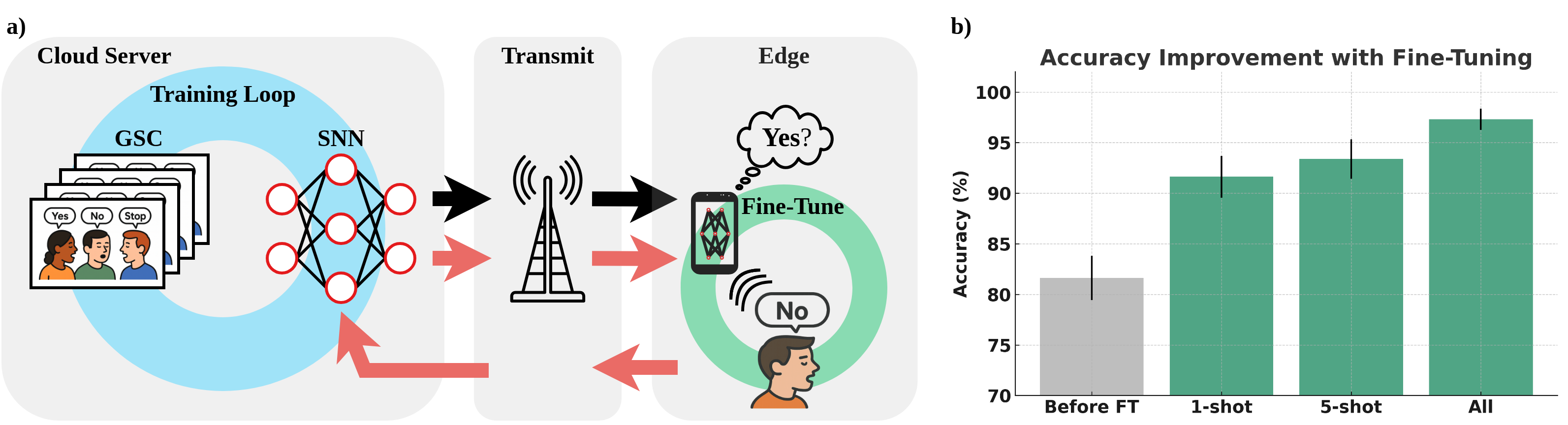}
       \caption{Google Speech Commands on-device fine-tuning (FT) with Traces Propagation. \textit{a)} Edge deployment pipeline for speech recognition. A neural network is pre-trained on the Google Speech Command dataset in the cloud (blue circle) and transmitted to the edge device (black arrows). Without fine-tuning, the model can misclassify inputs (e.g. mistaking a "yes" for a "no). If fine-tuning is not available at the edge, the new user recordings must be sent back to cloud (red arrows) for retraining, reducing privacy and increasing cost. With on-device fine-tuning (green circle) the model can adapt to the new user without data transmission. \textit{b)} Fine tuning accuracy improvement with 1-shot, 5-shot, and all user samples. Before FT represents the accuracy on GSC for the new user following cloud-training and deployment.}
    \label{fig:fine_tuning}
\end{figure}

In the previous section, we demonstrated the effectiveness of TP in training from scratch on standard SNN datasets. However, practical deployment on an edge device requires adaptation to new user distributions or environments rather than training from scratch. For example, in speech recognition tasks, the model is conventionally trained on a cloud server and subsequently deployed to an edge device, as depicted in Fig.~\ref{fig:fine_tuning}a. This process fails to account for user-specific variations, such as speaking style, accent, or ambient noise, reducing performance once deployed. Adapting the model to these variations requires collecting user-specific samples and transmitting them to the cloud for retraining, introducing challenges related to privacy, latency, and bandwidth usage. 

Motivated by the efficient and fully local update rule of TP, we investigated its applicability for on-device fine-tuning using the Google Speech Commands (GSC) dataset. The details of the dataset pre-processing and split are discussed in detail in Section~\ref{sec:material_and_method:setup_and_datasets}. With this objective, we trained an SNN of two hidden layers with recurrent connections on all user recordings of the GSC dataset, except those of the user \textit{c50f55b8}, which we reserved to simulate on-device fine-tuning. As can be seen from the gray bar in Figure~\ref{fig:fine_tuning}b, which represents the accuracy of the model on the excluded user following deployment, the model correctly classifies only 81.62 \% of its samples, highlighting the need for user-specific adaptation. To evaluate the capabilities of TP to perform fine-tuning, we retrained the model using a limited number of samples for 3 epochs. Specifically, we perform 1-shot, 5-shot and all-shot fine-tuning, where only 1, 5, or all available recordings per class of the excluded user are used for retraining the model. As shown in Figure~\ref{fig:fine_tuning}b, TP achieves significant improvement in accuracy, providing +10 pp, +11.76 pp and +15.68 pp for 1-shot, 5-shot, and all-shot, respectively.

Furthermore, we evaluated the ability of the model to learn solely from the user-specific samples without pre-training on other user data. In this case, performance remained low across all cases ($10.14 \pm 4.73\%$ for 1-shot, $12.84 \pm 0.68\%$ for 5-shot, and $12.84 \pm 4.73\%$ for all-shot), highlighting the importance of initial pre-training. In addition, we measured the amount of performance drop on the pre-trained model after fine-tuning, to measure the level of \textit{catastrophic forgetting}.  Specifically, the test accuracy decreased from $79.67 \pm 0.07\%$ before fine-tuning to $74.90 \pm 0.25\%$, $73.97 \pm 0.36\%$, and $74.75 \pm 0.32\%$ for 1-shot, 5-shot, and all-shot settings, respectively.

\section{Materials and Methods}
\label{sec:material_and_methods}
\subsection{Setup and Datasets}
\label{sec:material_and_method:setup_and_datasets}
In this section, we describe the datasets used in our experiments, together with the pre-processing steps applied to each. For each dataset, we performed hyperparameter tuning on the membrane potential decay constant $\alpha$, the trace integration constant $\beta$, and the threshold $v_{th}$, which was selected from the values 0.5 and 1.0. All different architectures used in our experiment use a simple integrator as an output layer, where the predicted class corresponds to the neuron with the highest integration value at the end of the sequence. For all datasets, we employ a learning rate of $1e^{-4}$ and apply the layer update rule described in Equation~\eqref{eq:local_grad} at each time step. As a surrogate function $\theta'$, we employ the ArcTan function defined in \cite{2021_fang} with a scale factor of 1. Following the hyperparameter tuning strategy, we train the network for 10 random seeds. Similarly to other works~\cite{ortner_2023, quintana_2024, apolinario_2025}, and report the best test accuracy encounter across the whole training process. Specifically, we report the mean and standard deviation of the peak test accuracy across the top 5 seeds. We also report the architectural configurations used in each case. For all convolutional architectures, a batch size of 64 is used, while for all fully connected architectures, a batch size of 128 is used. All experiments and implementation details can be found in the official repository\footnote{\url{https://github.com/lollopes/traces_propagation}} of this work.

\subsubsection{N-MNIST}

This dataset is the neuromorphic version of the conventional MNIST dataset used to benchmark ANNs. It is created by recording the MNIST digits with a Dynamic Vision Sensor (DVS) moved in three different saccades to mimic the behavior of the human eye. Similarly to \cite{quintana_2024}, we retain only the first saccade of the recording and convert the event streams into frame-based tensors of 100~ms with a time step of 1~ms. From an architectural standpoint, we used a single hidden layer of 200 neurons without recurrent connection. We find the best hyperparameters to be $\beta=0.98$, $\alpha=0.98$, and $v_{th}=1.0$.

\subsubsection{SHD}
The Spiking Heidelberg Digits (SHD) dataset contains recordings of spoken digits from 1 to 10, in both English and German, converted into spike trains using a biologically inspired cochlear model. The objective of the dataset is to classify the digit that is spoken. Unlike N-MNIST, this dataset is characterized by a strong temporal structure, which makes it particularly suitable for evaluating the ability of learning algorithms to assign credit over time in spiking neural networks. To convert the events stream into frame-based tensors, we accumulate the events in a time window of 10~ms and use the entire recoding duration. From an architectural standpoint, we used a single hidden layer of 450 neurons, with and without recurrent connections. We find the best hyperparameters to be $\beta=0.96$, $\alpha=0.97$, and $v_{th}=1.0$, for the feed-forward architecture, and $\beta=0.85$, $\alpha=0.85$, and $v_{th}=0.5$ for the recurrent one. 

\subsubsection{DVS-GESTURE}
This dataset consists of 11 hand gestures from 29 subjects under 3 illumination conditions recorded using a DVS camera. The goal of this task is to correctly classify each of the 11 hand gestures. We follow the same pre-processing steps as in \cite{apolinario_2025}. Specifically, sequences of different lengths are split into samples of 1.5 seconds, and events are accumulated using a time window of 75~ms, leading to frame tensors of 20 time steps. Furthermore, each frame is resized to 32 $\times$ 32 and randomly cropped with a zero-padding of 4. From an architectural stand point, we employ the same VGG-9 structure as~\cite{apolinario_2025}, to guarantee a fair comparison. Specifically, the architecture consists of 8 LIF based convolutional layers with 3 $\times$ 3 kernels, where the number of channels increases from 64 to 512, and max pooling is applied every 2 players. To stabilize the firing behavior across the whole network, weight normalization is applied to the input currents of each layer. We find the best hyperparameters to be $\beta=0.53$, $\alpha=0.5$, and $v_{th}=1.0$.

\subsubsection{DVS-CIFAR10}
This dataset is the neuromorphic version of the conventional CIFAR10 used to benchmark CNNs. The conversion from static images to event-based streams was achieved by repeated closed-loop movement of the static image in front of a stationary DVS camera. Like in~\cite{apolinario_2025}, events are accumulated in 10 times steps and resized to 48 $\times$ 48. For the training set, the images are randomly cropped with zero-padding of 4. The same VGG-9 architecture explained in the previous section is used for this dataset. We find the best hyperparameters to be $\beta=0.18$, $\alpha=0.19$, and $v_{th}=0.5$.

\subsubsection{GSC}
The Google Speech Command (GSC) dataset is a common benchmark for keyword spotting. It comprises 65,000 one-second WAV audio clips of 30 short English words. To convert WAV audio recordings into a spike-based signal, each audio recording is sampled at 16~kHz and padded with zero if its duration is shorter than 1~s. The resulting signal is then segmented into overlapping windows of 30~ms with a hop of 10~ms. To each window a Fast Fourier Transform (FFT) is applied to extract its frequency components, which are then projected onto 40 mel-frequency bands distributed between 20~Hz and 4~kHz, to  mimic the human auditory perception. This processing generates a time-frequency representation of 101 frames per recording. Additionally, delta and delta-delta features are computed and added as extra channels. Finally, the resulting tensors are normalized and comprise 3 channels, 101 time steps, and 40 features. 

For the fine-tuning experiments outlined in Section~\ref{sec:result:use_cases:gsc_fine_tune}, we generate three sets: a training set, used to simulate cloud learning, multiple $k$-shot support sets for user-specific adaptation, and a query set used to evaluate the performance on the user after fine-tuning. Specifically, the training set comprises all samples from the training split of the original GSC dataset, excluding those of the user \textit{c50f55b8}, who has the most number of recordings. The samples of this user are then split with an 80-20 ratio into support and query sets, respectively, resulting in 242 support samples and 74 query samples. From the support set, we create multiple $k$-shot subsets by sampling $k$ examples per class, where $k \in \{1,5,242\}$. These $k$-shot support sets are used to perform the fine tuning on user-specific data, while the query set is used to evaluate the performance of the tuned model to unseen words of the user. 

For our experiment, we train an SNN with one hidden layer of 450 neurons with recurrent connection followed by a linear integrator as the classification layer using the TP learning rule. To find the best hyperparameter we perform a hyperparameter optimization and found the best parameters to be $v_{th}=0.66$, $\beta=0.98$, and $\alpha=0.77$. We train on the samples from the train set for 8 epochs using a batch size of 1024, and perform the fine-tuning stages for the different $k$-shots for only 3 epochs, to emulate on-device adaptation.

\subsection{Characterization of Trace Propagation}
\label{sec:material_and_method:characterization_tp}

\begin{figure}[t]
    \includegraphics[width=\textwidth]{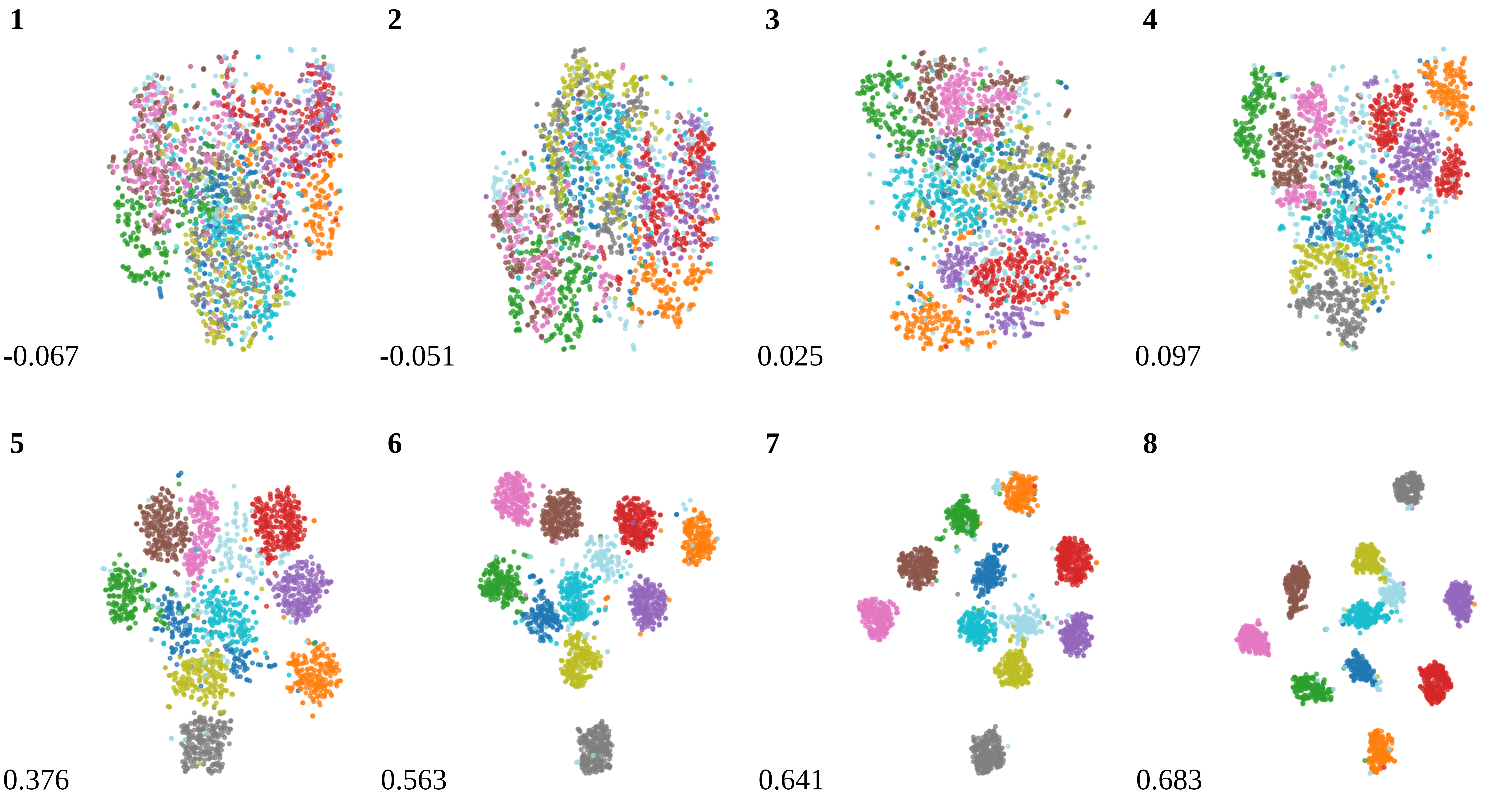}
    \caption{t-SNE visualization of input traces across layers on the DVS-GESTURE dataset. Panels 1 to 8 show the 2D representation at the end of training of the input traces for each layer of a VGG-9 architecture trained on the DVS-GESTURE dataset. In the bottom-left corner of each panel, we report the Silhouette score~\cite{shahapure_2020} to quantify the degree of clustering. A value of -1 indicates overlapping or poorly separated clusters, while a value of +1 correspond to more compact and well-separated groups. As we can see, the Silhouette score increases progressively with layer depth, demonstrating that the network develops increasingly discriminative and separable representations.}
    \label{fig:tsne}
\end{figure}

As explained in Section~\ref{sec:result:traces_propagation}, the main objective of TP is to group the input traces into class-specific clusters. To validate this operating principle and provide an intuitive visualization of this effect, we provide a t-SNE~\cite{maaten_2008} visualization of the input traces at the end of training for each convolutional layer of the VGG-9 architecture trained on the DVS-GESTURE dataset. Specifically, we perform the t-SNE analysis using the last time step values of the input traces, as they represent a summary of the neuron activity across the whole sequence that contributed to the network's final classification. Specifically, we collect the traces of each example across the last epoch and perform a t-SNE analysis at each layer sampling only 2000 examples. The results of this analysis are shown in Figure~\ref{fig:tsne}. From this figure, it can be observed that in the first two convolutional layers (panels 1 and 2), the 2D representations of the traces for the 11 classes are still mixed together. A clearer separation emerges in the third and fourth layers (panels 3 and 4). While distinct clusters emerges, they radius of the cluster remains significantly large, with some overlap between classes. On the other hand, for layers 5 to 8, the traces start to reduce their radius and are pushed further apart in the feature space, indicating linear separability between classes.


\subsection{MACs and memory estimation for TESS and TP}
\label{sec:material_and_method:mac_and_mem}
In this section, we explain the methodology used to estimate the total memory cost and the total Multiply-Accumulate Operations (MACs) for both TESS and TP, as reported in Table~\ref{tab:convolutional}. The total MACs of TP are estimated as follows:  

\begin{equation}
\mathrm{MACs}_{\mathrm{TP}} 
= T \sum_{l=0}^{L} \left[
      \underbrace{2B H_l}_{\eqref{eq:trace_in} \& \eqref{eq:trace_trg} }
    + \underbrace{B^{2} H_l}_{\eqref{eq:act_comp}}
    + \underbrace{B^{2} H_{l-1}}_{\eqref{eq:act_comp_prev}}
    + \underbrace{2B^{2} H_{l-1} H_l}_{\eqref{eq:local_grad}}
\right].
\end{equation}

In \cite{apolinario_2025}, the authors provide an estimate of the MACs, but consider only the modulation signal, leaving out the computational cost of the traces. Here, we want to fully characterize the computational cost and include all necessary signals. To estimate the total MACs we refer to the equations numbers of the original work~\cite{apolinario_2025}. The total MACs for TESS can be estimated according to Equation \eqref{eq:tess_macs}, where $t_l$ represents the time-step at which the update is performed.

\begin{equation}
\label{eq:tess_macs}
\mathrm{MACs}_{\mathrm{TESS}} 
= (T-t_l) \sum_{l=0}^{L} \left[
    \underbrace{BH_l}_{(7)} +
    \underbrace{BH_{l-1}}_{(5)} +
    \underbrace{BH_lO + BH_l}_{(9)} + 
    \underbrace{2B H_{l-1} H_l}_{(11) \& (12)} \right] 
\end{equation}

\begin{figure}[t!]
    \centering
    \includegraphics[width=0.9\textwidth]{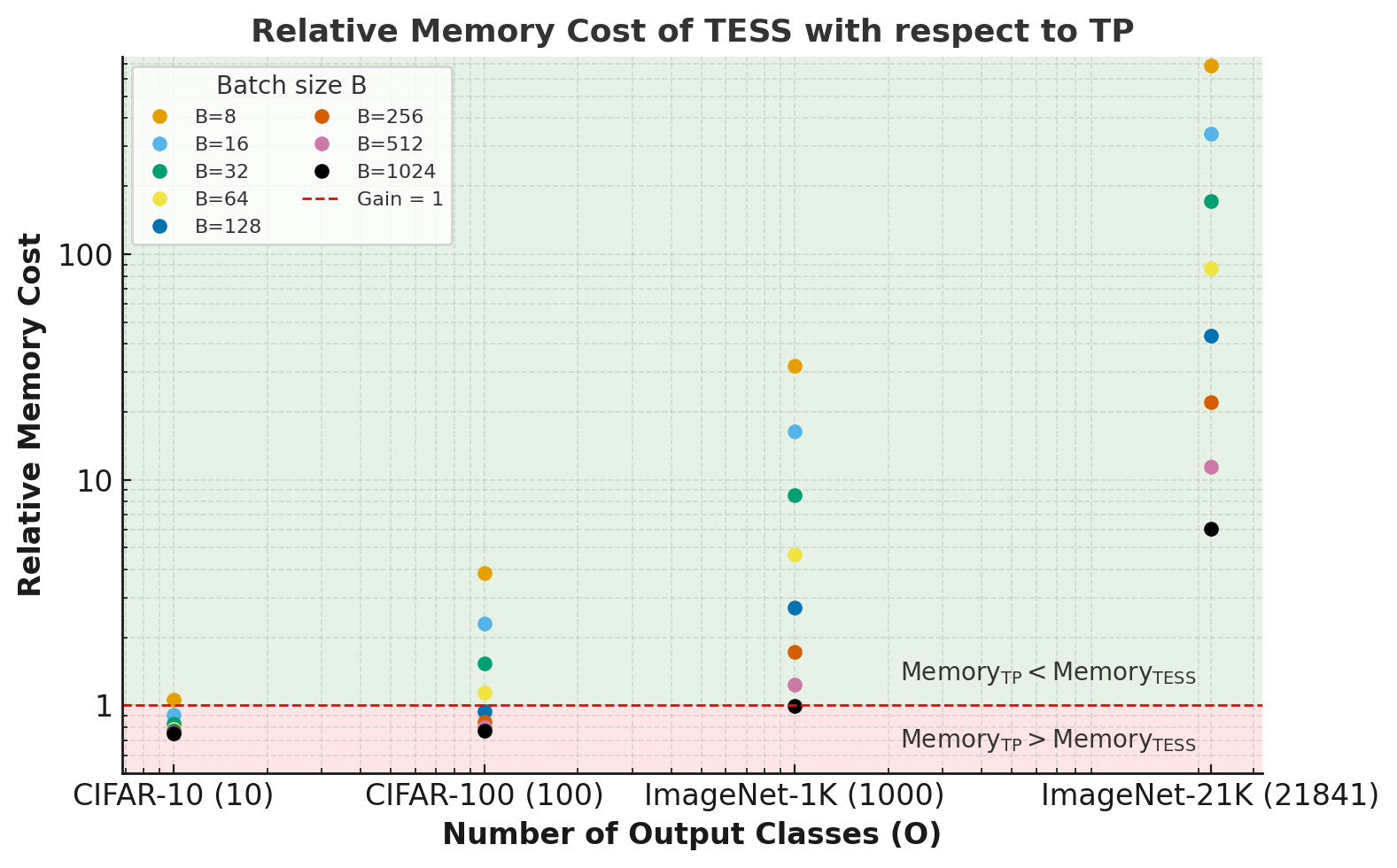}
    \caption{Relative memory cost of TP compared to TESS as a function of the number of output classes and batch size. A relative memory cost of $G$ indicates that TESS uses $G$ times more memory than TP.}
    \label{fig:rmc}
\end{figure}

On the other hand, the memory estimations for TESS and TP are calculated according to Equation~\eqref{eq:tp_mem} and~\eqref{eq:tess_mem}, respectively. 

\begin{align}
\mathrm{Memory}_{\mathrm{TP}} &=                       
    \underbrace{OH_0}_{\textit{Auxiliary matrix S}} 
    +\sum_{l=0}^{L} 
    \left[\underbrace{2BH_l}_{\textit{input \& target potential}}
    + \underbrace{2BH_l}_{\textit{input \& target trace}}\right]\label{eq:tp_mem}
\\
    &= OH_0 + 4B\sum_{l=0}^{L}H_l \label{eq:tp_mem_1}
\end{align}

\begin{align}
\label{eq:tess_mem}
\mathrm{Memory}_{\mathrm{TESS}} &=  \sum_{l=0}^{L} \left[
     \underbrace{BH_l}_{\textit{input potential}}
    + \underbrace{2BH_l}_{\textit{traces}} +
     \underbrace{OH_l}_{\textit{Auxiliary matrices } B_l}\right] \\
     &= (3B+O)  \sum_{l=0}^{L}H_l
\end{align}

For deep networks, the term $\sum_{l=0}^{L} H_l$ dominates $H_0$. Thus, we can approximate \eqref{eq:tp_mem_1} as $ 4B\sum_{l=0}^{L}H_l$. Taking the ratio of memory estimations for TESS and TP under this assumption, we express the relative memory cost of TESS with respect to TP as shown in Equation~\ref{eq:gain_mem}. From this equation, we can deduce that TP is more memory efficient whenever the number of output classes is greater than the batch size ($O > B$). For complex datasets, where the number of output classes dominates the batch size ($O >> B$), TP provides a memory gain that scales with $O/4B$, indicating that the memory advantage of TP over TESS increases linearly with the number of output classes.

\begin{align}
\text{Relative Memory Cost}= \frac{\mathrm{Memory}_{\mathrm{TESS}}}{\mathrm{Memory}_{\mathrm{TP}}} \approx \frac{3B + O}{4B}.
\label{eq:gain_mem}
\end{align}

To provide an intuitive understanding of how the memory of TESS scales compared to that of TP, we plot Equation~\ref{eq:gain_mem} in Figure~\ref{fig:rmc} for conventional batch sizes and a set of widely-used image classification datasets, arranged in order of increasing number of output classes: CIFAR-10, CIFAR-100, ImageNet-1K, and ImageNet-21K.





\section{Discussions and Conclusions}
\label{sec:discuss_and_conlude}
The on-device adaptation required to deploy deep spiking neural networks in real-world use cases, where the environment dynamics are constantly changing, is challenged by the spatial and temporal non-locality of BPTT. To address this challenge, we propose Traces Propagation, a forward-only, scalable learning rule that is local in time and space. It combines the bio-inspired mechanism of eligibility traces for local temporal credit assignment, with a layer-wise contrastive loss inspired by the Signal Propagation algorithm to address the spatial credit assignment problem. 

Unlike previous scalable solutions such as TESS, TP removes the need for layer-wise auxiliary matrices at the cost of doubling the internal states, providing a favorable memory cost compared to TESS when $O>B$. In particular, TP provides a memory advantage over TESS that increases linearly with the number of output classes, while maintaining a space complexity of $\mathcal{O}(LH)$ as other state-of-the-art local learning rules. From a computational perspective, TP scales with a complexity of $\mathcal{O}(B^2LH)$, removing the dependency on the number of output classes $O$ present in other solutions, making it advantageous for complex datasets where $O \gg B$ (i.e. the number of output classes is much greater than the batch size). 

Across standard neuromorphic benchmarks, TP achieves state-of-the-art results on the N-MNIST and SHD dataset compared to other fully-local methods, with only $1.4$ percentage point loss compared to BPTT. In deep architectures such as VGG-9, TP achieves a performance comparable to that of TESS and BPTT on DVS-GESTURE, falling short of only $0.33$ pp, while incurring a loss of $3.38$ pp relative to TESS on DVS-CIFAR10. These results demonstrate that the proposed approach is sufficient to solve complex spatio-temporal tasks using deep architectures but can be hindered by datasets with high inter-class overlap, such as DVS-CIFAR10~\cite{2016_murthy}. Furthermore, we showcased the applicability of TP to practical use cases, such as fine-tuning on the Google Speech Commands dataset, where TP improved user-specific accuracy by up to $+11.76$, $+15.68$, and $+15.68$ percentage points for 1-shot, 5-shot, and all-shot settings, respectively.


Although TP provides competitive performance through a learning rule that is local in space and time, showcasing scalability to deep spiking architectures across datasets of varying complexity and use cases, several limitations remain. First, the contrastive nature of TP prevents weight updates to occur in an online per-sample fashion, making it advantageous to other solutions only when the batch size is smaller compared to the number of output classes. Second, the calculation of pairwise similarities between all examples in a give batch, introduces a computational cost of $B^2LH$, which can hinder its applicability in resource constrained devices for large batch sizes. Third, we suspect that the contrastive nature of TP limits its performance in datasets with substantial inter-class similarity, such as CIFAR-10~\cite{2016_murthy}, from which the DVS version is derived. 

Future extensions of this work include the investigation of the performance of TP in quantization-aware training to enable lightweight training on embedded devices, implementing TP on conventional SNN accelerators, and extending it to continuous learning settings where new output classes can be learned directly after deployment. Together, these directions highlight TP as a promising step toward enabling memory-efficient, adaptive, and scalable spiking neural networks for real-world edge intelligence.

\section*{Acknowledgment}
This publication is funded in part by the project NL-ECO: Netherlands Initiative for Energy-Efficient Computing (with project number NWA. 1389.20.140) of the NWA research programme Research Along Routes by Consortia which is financed by the Dutch Research Council (NWO). This research was also partially funded by the Dutch Organization for Scientific Research (NWO) through the Self-Healing Neuromorphic Systems (SNS) project (KICH1.ST04.22.021).

\section*{References}
\bibliographystyle{IEEEtran}

\newpage
\appendix
\section{}
\label{appendix:1}
\begin{algorithm}
\caption{Trace Propagation Algorithm }
\label{alg:trace_propagation}
\begin{algorithmic}[1]

    \State \textcolor{gray}{\# Requires batch size $B \geq 2$ due to contrastive loss}
    \For{each time step \( t \)}
    
        \For{each sample \( b = 1, \dots, B \)} 
            \State \textcolor{gray}{\# Input and target trace update (per sample)}
            \State $\epsilon_0^t = \beta_0 \epsilon_0^{t-1} + s_0^t$
            \State $\tilde{\epsilon}_0^t = \beta_0 \tilde{\epsilon}_0^{t-1} + c$

            \For{each layer \( l = 1, \dots, L \)}

                \State \textcolor{gray}{\# Input propagation}
                \State $v^t_l = \alpha_l v^{t-1}_l + s^t_{l-1} W_l - s^{t-1}_l v_{l,th}$
                \State $s^t_l = \Theta (v^t_l - v_{l,th})$
                \State $\epsilon_l^t = \beta_l \epsilon_l^{t-1} + s^t_l$

                \If{$l == 1$}
                    \State $\tilde{s}^t_0 = c$
                    \State $W_l = S$
                \EndIf

                \State \textcolor{gray}{\# Target propagation}
                \State $\tilde{v}^t_l = \alpha_l \tilde{v}^{t-1}_l + \tilde{s}^t_{l-1} W_l - \tilde{s}^{t-1}_l v_{l,th}$
                \State $\tilde{s}^t_l = \Theta (\tilde{v}^t_l - v_{l,th})$
                \State $\tilde{\epsilon}_l^t = \beta_l \tilde{\epsilon}_l^{t-1} + \tilde{s}^t_l$

            \EndFor
        \EndFor

        \For{each layer \( l = 1, \dots, L \)}
            \State \textcolor{gray}{\# Contrastive similarity and synaptic update (batch-wise)}
            \State $z^t_l = \epsilon_l^t (\tilde{\epsilon}_l^t)^\top$
            \State $y^t_l = \mathrm{Softmax}\!\left(-f_{\mathrm{sim}}(\tilde{\epsilon}_{l-1}^t, \tilde{\epsilon}_{l-1}^t)\right)$

            \State $\Delta W^t_{l,\mathrm{in}} = (Softmax(z_l^t) - y_l^t) \, \tilde{\epsilon}_l^t \, \Theta'(v_l^t - V_{\text{th}}) \, s^t_{l-1}$
            \State $\Delta W^t_{l,\mathrm{trg}} = (Softmax(z_l^t) - y_l^t)\, \epsilon_l^t \, \Theta'(\tilde{v}_l^t - V_{\text{th}}) \, \tilde{s}^t_{l-1}$

            \State $W_l^{t+1} = W_l^t - \eta \left(\Delta W^t_{l,\mathrm{in}} + \Delta W^t_{l,\mathrm{trg}}\right)$
        \EndFor
    \EndFor
\end{algorithmic}
\end{algorithm}


\end{document}